\documentclass[11pt]{article}
\date{}

\usepackage[left=25mm, right=25mm, top=25mm, bottom=25mm, includehead=false, includefoot=false]{geometry}

\usepackage{makecell}
\usepackage{graphicx}
\usepackage{url}
\usepackage[round,semicolon]{natbib}  
\usepackage{authblk} 

\usepackage[table]{xcolor}
\usepackage[parfill]{parskip} 
\usepackage{amsmath}
\usepackage{sectsty}
\allsectionsfont{\sffamily}

\usepackage[pdftex]{hyperref} 
\hypersetup{pdfborder={0 0 0} }

\title{\sffamily\fontsize{16}{0}\textbf{Mapping Languages and Demographics \\ with Georeferenced Corpora}}

\author[1]{Jonathan Dunn}
\author[2]{Benjamin Adams}
\affil[1]{Department of Linguistics, University of Canterbury, Christchurch, New Zealand}
\affil[2]{Department of Geography, University of Canterbury, Christchurch, New Zealand}
\affil[1]{\texttt{Email: \url{jonathan.dunn@canterbury.ac.nz}}}

\begin{document}

\maketitle

\begin{abstract}
\noindent
\setlength{\parindent}{0pt}

This paper evaluates large georeferenced corpora, taken from both web-crawled and social media sources, against ground-truth population and language-census datasets. The goal is to determine (i) which dataset best represents population demographics; (ii) in what parts of the world the datasets are most representative of actual populations; and (iii) how to weight the datasets to provide more accurate representations of underlying populations. The paper finds that the two datasets represent very different populations and that they correlate with actual populations with values of $r=0.60$ (social media) and $r=0.49$ (web-crawled). Further, Twitter data makes better predictions about the inventory of languages used in each country.

$ $ 

\end{abstract}


\section{Introduction}

In recent years there has been an increasing amount of research investigating the use of unstructured geographic information, such as user-generated social media content and textual web data, for geographical analysis. Such unstructured data sources can provide information about places that is difficult to measure through traditional sensor observations. Aggregated representations of places that are extracted from textual corpora, e.g., give us insight into what people feel and think about places, potentially providing a much richer understanding of human geography \citep{rattenbury2009methods,adams2013inferring}. It can also give insight into the relationships between places and human behavior \citep{hu2017extracting,van2019go}. However, a recurring issue with this kind of big data, and user-generated content in general, is the question of how representative these data sets are compared to the populations that we wish to study \citep{miller2015data}. There exists little previous empirical work to establish how representative web corpora are with respect to different geographic regions of the world (c.f. \cite{hecht2014tale,malik2015population}). In this paper we describe such a computational experiment using language identification models on two global-scale corpora. 

How well does language data represent both regional population densities and the social characteristics of regional populations? To answer this question, we collect and analyze two large global-scale datasets: web-crawled data from the Common Crawl (16.65 billion words) and social media data from Twitter (4.14 billion words). This paper evaluates demographic-type information that is derived from these datasets, information which traditionally has been collected using survey-instruments as part of a census. 

When labeled with a language identification model, this data provides a representation of both (i) how much language a particular country produces, a proxy for population density and (ii) the mix of languages used in a country, a proxy for population demographics. These corpus-based representations are compared against four ground-truth baselines. First, the UN country-level population estimates \citep{UnitedNations2017}. Second, because not all populations have equal access to internet technologies, we use per capita GDP \citep{UnitedNations2017a} and internet-usage statistics \citep{UnitedNations2011} to adjust raw populations. Third, the UN country-level census aggregations are used to represent what languages are used in each country \citep{UnitedNations2018} and, where these are not available, the \textit{World Factbook} \citep{Agency2019} estimations are used. The goal is to measure how well corpus-based representations correspond with each of these ground-truth, survey-based representations. Thus, we are not concerned at this point if the corpus-based representations are skewed or inaccurate in particular locations. Rather, the purpose is to measure how and where these datasets are skewed as a method for evaluating and improving future data collection methods.

We can view this problem from two perspectives: 1) from a human geography perspective, is it possible to use global-scale corpora to understand characteristics of regional populations?, and 2) from the perspective of computational linguistics, is it possible to normalize corpora to proportionally represent diverse populations? For example, some countries (like the United States) and some languages (like English) dominate many datasets. Is it possible to systematically correct this imbalance?

We begin by describing the corpora and how they were collected (Section 2) and the language identification model that is used to label them with language codes (Section 3). After looking at the frequency distribution of languages across the entire dataset (Section 4), we undertake a country-level evaluation of the datasets, first against population-density baselines (Section 5) and then against language-use baselines (Section 6).

\section{Collecting Global Corpora}

Data comes from two sources of digital texts: web pages from the Common Crawl\footnote{\url{http://www.commoncrawl.org}} and social media from Twitter\footnote{\url{http://www.twitter.com}}. Starting with the web-crawled data, we can compare this dataset to previous georeferenced web corpora \citep{geq12,df15}. The basic pipeline is to process all text within $<p>$ tags, removing boilerplate content, navigation content, and noisy text. We view each web page as a document containing the remaining material. Documents are then deduplicated by site, by time, and by location.\footnote{\url{https://github.com/jonathandunn/common_crawl_corpus}}

\begin{table}
\begin{center}
	\begin{tabular}{|l|cc|cc|cc|}
		\hline
		\textbf{Region} & \makecell{\textbf{Population} \\ \textbf{(By \%)}} 
		& \makecell{\textbf{N.} \\ \textbf{Cntry}} & \makecell{\textbf{Web} \\ \textbf{(By \%)}} & \makecell{\textbf{N.} \\ \textbf{Cntry}} & \makecell{\textbf{Twitter} \\ \textbf{(By \%)}} & \makecell{\textbf{N.} \\ \textbf{Cntry}} \\
		\hline
		Africa, North & 3.4\% & 9 & 0.7\% & 6 & 2.1\% & 9 \\
		Africa, Southern & 1.0\% & 4 & 0.4\% & 3 & 2.1\% & 3 \\
		Africa, Sub-Saharan & 10.1\% & 33 & 2.6\% & 25 & 6.1\% & 29 \\
		\hline
		America, Brazil & 2.8\% & 1 & 1.3\% & 1 & 2.9\% & 1 \\
		America, Central & 2.9\% & 25 & 5.3\% & 24 & 9.3\% & 15 \\
		America, North & 4.8\% & 2 & 1.4\% & 2 & 8.5\% & 2 \\
		America, South & 2.9\% & 11 & 7.0\% & 11 & 9.7\% & 11 \\
		\hline
		Asia, Central & 2.7\% & 10 & 5.8\% & 9 & 2.5\% & 9 \\
		Asia, East & 22.3\% & 8 & 13.2\% & 7 & 2.3\% & 6 \\
		Asia, South & 23.3\% & 7 & 2.7\% & 5 & 8.0\% & 7 \\
		Asia, Southeast & 8.4\% & 22 & 12.1\% & 15 & 5.9\% & 16 \\
		\hline
		Europe, East & 2.4\% & 17 & 27.4\% & 17 & 7.8\% & 17 \\
		Europe, Russia & 2.0\% & 1 & 0.6\% & 1 & 2.5\% & 1 \\
		Europe, West & 5.7\% & 25 & 14.6\% & 24 & 19.9\% & 23 \\
		\hline
		Middle East & 4.5\% & 15 & 4.0\% & 11 & 5.4\% & 13 \\
		\hline
		Oceania & 1.0\% & 8 & 1.0\% & 4 & 5.1\% & 7 \\
		\hline
		\textsc{\textbf{Total}} & \makecell{\textbf{7.35 billion} \\ \textbf{people}} & \textbf{199} & \makecell{\textbf{16.65 billion} \\ \textbf{words}} & \textbf{166} & \makecell{\textbf{4.14 billion} \\ \textbf{words}} & \textbf{169} \\
		\hline
	\end{tabular}
	\caption{Relative Size of Georeferenced Corpora by Region with Population Baseline}
	\label{tab:1}
	\end{center}
\end{table}

Language samples are geo-located using country-specific top-level domains: the assumption is that a language sample from a web-site under the \textit{.ca} domain originated from Canada. This approach to regionalization does not assume that whoever produced that language sample was born in Canada or represents a traditional Canadian dialect group. Rather, the assumption is only that the sample represents someone in Canada who is producing language data. Previous work has shown that there is a significant relationship between domain-level georeferencing and traditionally-collected linguistic data \citep{Cook2017}.

Some countries are not available because their top-level domains are used for other purposes (i.e., \textit{.ai}, \textit{.fm}, \textit{.io}, \textit{.ly}, \textit{.ag}, \textit{.tv}). Domains that do not contain geographic information are also removed from consideration (e.g., \textit{.com} sites). The Common Crawl dataset covers 2014 through the end of 2017, totalling 81.5 billion web pages. As shown in Table 1, after processing this produces a corpus of 16.65 billion words.\footnote{\url{https://labbcat.canterbury.ac.nz/download/?jonathandunn/CGLU_v3}} Table 1 also shows the number of countries represented in the web corpus against the number of countries in the ground-truth UN dataset and in the collected Twitter corpus. Countries may be missing from the web dataset (i) because their domains are used for a different purpose or (ii) their domains are not widely used or the country does not produce a significant amount of data on the open internet.

In isolation, web-crawled data provides one observation of global language use. Another common source of data used for this purpose is Twitter \citep{eosx10,v12,kcdsbhskv13,mbpgzv13,eosx14,ghg14,ds17}). A spatial search is used to collect tweets from within a 50km radius of 10k cities.\footnote{\url{https://github.com/datasets/world-cities}} This search avoids biasing the selection by using language-specific keywords or hashtags. The Twitter data covers the period from May of 2017 until early 2019. This creates a corpus containing 1,066,038,000 tweets, all connected with the city from which they were collected. Because the language identification component only provides reliable predictions for samples containing at least 50 characters, the corpus is pruned to that length threshold (this removes approximately 250 million tweets). As shown in Table 1, this produces a corpus containing 4.14 billion words.

Each of these datasets rests on different assumptions and, as a result, is subject to different confounds. For example, using top-level domains to determine the country of origin for web pages is likely to over-represent countries like Estonia that use their TLD extensively and to under-represent countries like the United States that do not traditionally use their TLD. The Twitter dataset relies on georeferenced tweets, but not all users have GPS-enabled devices. For example, we might expect that countries with a lower per capita GDP have a lower percent of georeferenced tweets, in addition to having fewer tweets overall. The goal here is to establish a baseline of how well these corpora represent actual populations.

\section{Language Identification}

Language identification (LID) is important because we need to identify as many languages as possible while using the small samples provided by social media. We use a multi-layer perceptron with character trigrams, trained using samples of 50 characters from a range of datasets. The focus here is to evaluate the performance of the LID component within the context of language mapping.\footnote{LID Code Download: \url{https://github.com/jonathandunn/idNet}} Previous work on LID either required large, homogenous samples \citep{Brown2014} or covered only about one hundred languages \citep{Liu2011,Liu2012}. The goal here is to evaluate the LID component on as many samples as possible from as many domains as possible. The model covers 464 languages drawn from the datasets represented in Table 2; this can be seen as a compromise between the sample size required and the number of languages covered.\footnote{LID Model Download: \url{https://labbcat.canterbury.ac.nz/download/?jonathandunn/idNet_models}} Table 2 shows the performance of the model on a held-out test set; the number of languages for each section of the test set is shown (listed as \textit{N. Langs}) as well as the number of unique test samples for that section (listed as \textit{Test Samples}). The model covers 464 languages because, after enforcing a threshold of the number of samples required for the development/training/testing datsets, this is how many languages remain. It is a purely practical number that is limited by the datasets that are drawn from to create the LID model. On the other hand, only 205 (web-crawled) and 97 (Twitter) languages were present with at least 100k words; thus additional languages are unlikely to be informative.

\begin{table}
	\centering
	\begin{tabular}{|l|c|ccc|c|}
		\hline
		\textbf{Dataset} & \textbf{N. Langs} & \textbf{Precision} & \textbf{Recall} & \textbf{F-Measure} & \textbf{Test Samples} \\
		\hline
		Bibles & 85 & 0.98 & 0.98 & 0.98 & 76,611 \\
		LTI (Bibles and UN) & 428 & 0.98 & 0.97 & 0.98 & 421,165 \\
		Tanzil & 38 & 1.00 & 1.00 & 1.00 & 38,201 \\
		\hline
		Europarl & 21 & 1.00 & 1.00 & 1.00 & 21,109 \\
		United Nations & 6 & 1.00 & 0.99 & 1.00 & 6,060 \\
		JRC & 21 & 0.97 & 0.96 & 0.96 & 21,008 \\
		\hline
		EU Books & 25 & 0.99 & 0.98 & 0.98 & 24,442 \\
		GlobalVoices & 31 & 0.99 & 0.98 & 0.98 & 28,565 \\
		NewsCommentary & 10 & 1.00 & 0.99 & 1.00 & 10,075 \\
		Wikipedia & 87 & 0.96 & 0.05 & 0.95 & 87,431 \\
		Setimes & 8 & 1.00 & 0.99 & 1.00 & 8,080 \\
		\hline
		Gnome & 74 & 0.97 & 0.96 & 0.96 & 72,257 \\
		Ubuntu & 71 & 0.97 & 0.96 & 0.96 & 66,673 \\
		\hline
		OpenSubtitles & 45 & 0.98 & 0.98 & 0.98 & 45,450 \\
		Tatoeba Sent. & 37 & 0.99 & 0.99 & 0.99 & 34,834 \\
		TED Talks & 52 & 0.99 & 0.99 & 0.99 & 49,472 \\
		\hline
		IARPA Babel & 11 & 1.00 & 0.99 & 0.99 & 11,016 \\
		Emille & 6 & 0.86 & 0.84 & 0.83 & 6,060 \\
		Indian Parallel & 6 & 1.00 & 0.99 & 1.00 & 6,060 \\
		\hline
		Twitter (Over 50)* & 25 & 0.98 & 0.94 & 0.96 & 23,791 \\
		\hline
	\end{tabular}
	\caption{Language Identification Performance Across Domains for Samples of 50 characters}
	\label{tab:1}
\end{table}

The dataset used for training and evaluating the LID component contains several independent sources of data: The first, more formal, set of domains comes from a traditional LID source: religious texts. Bibles are taken from \citep{Christodouloupoulos2015} and from \citep{Brown2014} (this is listed as \textit{LTI} in Table 2); translations of the Quran are taken from the Tanzil corpus \citep{Tiedemann2012}. The second set of domains contains official government and legislative texts: the European parliament, the JRC-Acquis corpus of European Union texts, and the United Nations \citep{Tiedemann2012}. The third set contains non-official formal texts: the EU Bookshop corpus \citep{Skadins2014}, newspapers and commentary from GlobalVoices, NewsCommentary, and Setimes \citep{Tiedemann2012}, and Wikipedia \citep{m12}. The fourth set contains documentation from open source software packages: Ubuntu and Gnome \citep{Tiedemann2012}. The fifth set mimics informal speech: OpenSubtitles covering movies and television, TED Talks \citep{Tiedemann2012}, and Tatoeba for language-learning sentences (from tatoeba.org). The sixth set contains language-focused corpora collected to represent specific languages: the Emille corpus of Indian languages \citep{Baker2004}, the Indian Parallel Corpus \citep{Post2012}, and the IARPA Babel project language packs, for example the Cantonese corpus \citep{Andrus2016}.

The official Twitter LID data \citep{Twitter2015} is also used for the evaluation (note that not all samples from the original dataset are still available). Given the length constraints of the model, this considers only samples containing at least 50 characters after cleaning has been performed. These results, with an F1 of 0.96, show that the LID model can also be used on tweets containing at least 50 characters. It is important to note that the LID model is trained using samples of 50 characters and that no Twitter data was included in the training set. Thus, this result represents the case of Twitter being an out-of-sample domain. It may be the case that future work could produce a LID model capable of accurate predictions on tweets with less than 50 characters. The present model, however, has been trained and evaluated using samples of 50 characters.

Table 2 shows the F1 score of a single LID model that is evaluated on held-out test samples of 50 characters from each domain. This reflects the expected accuracy of the language labels applied to the types of data found in the web-crawled and social media datasets. These datasets are dominated by more widely used languages: only 205 languages are present with at least 100k words in the web-crawled dataset and only 97 in the social media dataset. This means that small minority languages are less likely to be represented here. This fixed threshold of 100k per language is a somewhat arbitrary limit; future work will consider the relative usage of a language by place (i.e., a threshold such as 5\% of the language produced by a country) to avoid a geographic bias against non-Western languages.

\section{Language Distribution}

\begin{table}
	\centering
	\begin{tabular}{|lcrcr|lcrcr|}
		\hline
		\textbf{Lang.} & \textbf{CC} & \textbf{CC\%} & \textbf{TW} & \textbf{TW\%} & \textbf{Lang.} & \textbf{CC} & \textbf{CC\%} & \textbf{TW} & \textbf{TW\%} \\
		\hline
		English & 1 & 29.9\% & 1 & 37.4\% & Bulgarian & 11 & 2.1\% & 39 & 0.1\% \\
		Spanish & 2 & 10.5\% & 2 & 19.0\% & Arabic & 12 & 2.0\% & 4 & 4.3\% \\
		Russian &  3 & 8.1\% & 6 & 3.0\% & Indonesian & 13 & 1.9\% & 7 & 2.5\% \\
		Serbo-Croatian & 4 & 6.7\% & 18 & 0.8\% & Latvian & 14 & 1.9\% & 37 & 0.1\% \\
		Mandarin & 5 & 5.0\% & 44 & 0.1\%  & Estonian & 15 & 1.7\% & 43 & 0.1\% \\
		German & 6 & 3.0\% & 9 & 1.7\% & Slovak & 16 & 1.3\% & 69 & 0.0\% \\
		French & 7 & 2.8\% & 5 & 4.2\% & Azerbaijani & 17 & 1.0\% & 40 & 0.1\% \\
		Slovenian & 8 & 2.6\% & 34 & 0.2\% & Romanian & 18 & 1.0\% & 31 & 0.3\% \\
        Portuguese & 9 & 2.5\% & 3 & 4.8\% & Icelandic & 19 & 0.8\% & 55 & 0.0\% \\
        Finnish & 10 & 2.3\% & 24 & 0.4\% & Italian & 20 & 0.8\% & 11 & 1.3\% \\
		\hline
		\textbf{These 10} & ~ & \textbf{73.4\%} & ~ & \textbf{71.6\%} & \textbf{These 20} & ~ & \textbf{87.9\%} & ~ &  \textbf{80.4\%} \\
		\hline
	\end{tabular}
	\caption{Most Common Languages by Frequency Rank and Percent of Corpora}
	\label{tab:1}
\end{table}

\textbf{To what degree do these datasets represent majority languages?} This is an important question because, with only language labels available, the prevalence of only a few languages will obscure important demographic information. Table 3 shows the top twenty languages (chosen from the web corpus) by their relative proportion of each dataset and, at the bottom, by their combined percent of the overall dataset. The two datasets do not agree in top languages given only the total number of words; however, these twenty languages make up a similar percent of each dataset.

We see that 87.9\% and 80.4\% of the data belongs to these twenty languages. The implication is that all the other languages make up less than 20\% of both datasets. This is potentially problematic because majority languages such as English and Spanish (both very common) are used across widely different demographics. In other words, knowing that a population uses English or Spanish gives us relatively little information about that population. A different view of this is shown in Figure 1, with the distribution by percentage of the data for the top 100 languages in each dataset (not necessarily the same languages). There is a long-tail of minority languages with a relatively small representation. This trend is more extreme in the social media dataset, but it is found with the same order of magnitude in both datasets. The figure is cut off above 2.0\% in order to visualize the long-tail of very infrequent languages. The biggest driver of this trend is English, accounting for 37.46\% of social media and 29.96\% of web data. This is the case even though both datasets have large numbers of observations from locations which are not traditionally identified as English-speaking countries, suggesting that in digital contexts these countries default to global languages which they do not use natively.

\begin{figure}
	\includegraphics[scale=0.99]{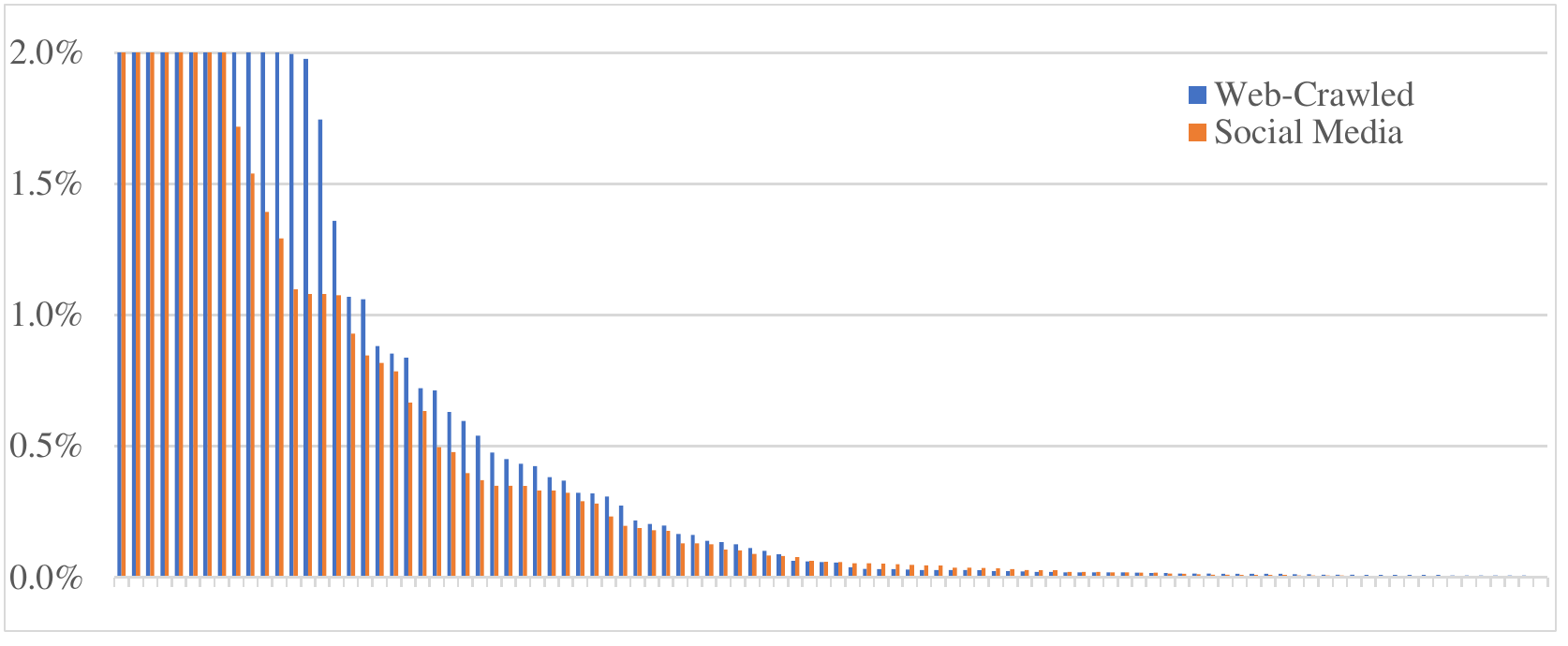}
	\caption{Distribution of Top 100 Languages By Percentage}
\end{figure}

Does this mean that digital data cannot be used to represent non-digital language use? The purpose of this paper is to find where and when and how we can map populations using digital data in order to establish a baseline for evaluating collection methods. The relative amount of data can be related to ground-truth population numbers (without language labels), and the language labels most common to particular countries can be related to ground-truth language-use statistics.

\section{Population Density}

\textbf{How well does the amount of data correspond with the population density of each country?} In this section we compare the number of words\footnote{Note that Chinese characters have been segmented} in each corpus with the UN population estimates. These datasets cover 199 countries in total, although the web-crawled data only represents 166 countries and the Twitter data only represents 169 countries.

The Pearson correlations between different measures per country are shown in Table 4. There are five measures per country: the size of the web-crawled and social media corpora in words, the UN population estimates, population adjusted by per capita GDP, and population adjusted by access to the internet. The purpose of these different weightings is to take into consideration the fact that some countries may produce more digital text per person than others.

\begin{table}
	\centering
	\begin{tabular}{|l|ccccc|}
		\hline
 ~ & Web (\textsc{words}) & Twitter (\textsc{words}) & Pop (\textsc{raw}) & Pop (\textsc{inter}) & Pop (\textsc{gdp}) \\
\hline
Web (\textsc{words}) & -- & 0.05 & 0.39 & 0.49 & 0.28 \\
Twitter (\textsc{words}) & 0.05 & -- & 0.39 & 0.42 & 0.60 \\
Pop (\textsc{raw}) & 0.39 & 0.39 & -- & 0.83 & 0.57 \\
Pop (\textsc{inter}) & 0.49 & 0.42 & 0.83 & -- & 0.84 \\
Pop (\textsc{gdp}) & 0.28 & 0.60 & 0.57 & 0.84 & -- \\
		\hline
	\end{tabular}
	\caption{Correlation Between Corpus Size and Population}
	\label{tab:1}
\end{table}

First, we notice that there is very little relationship between the two corpora ($r=0.05$). This is interesting in and of itself because, given the systematic attempt to find georeferenced texts, we would expect a fairly high correlation here. But this shows that the data comes from different places (c.f., Figures 2 and 3). Because the collection methods were designed with this purpose in mind, it is unlikely that this low correlation is caused by the methods themselves rather than reflecting the fact that these datasets represent different populations. In other words, this is a strong indication that web data and Twitter data represent two different populations regardless of variations in the collection methods employed in this study.

\begin{figure}
    \centering
	\includegraphics[width=1\textwidth]{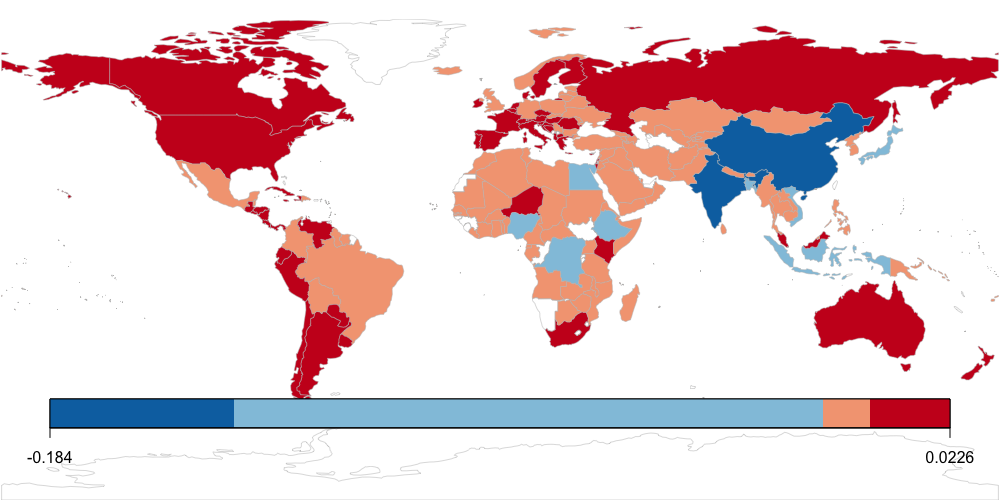}
	\caption{Representation of Twitter Data (Red = over-represented i.r.t population)}
\	\includegraphics[width=1\textwidth]{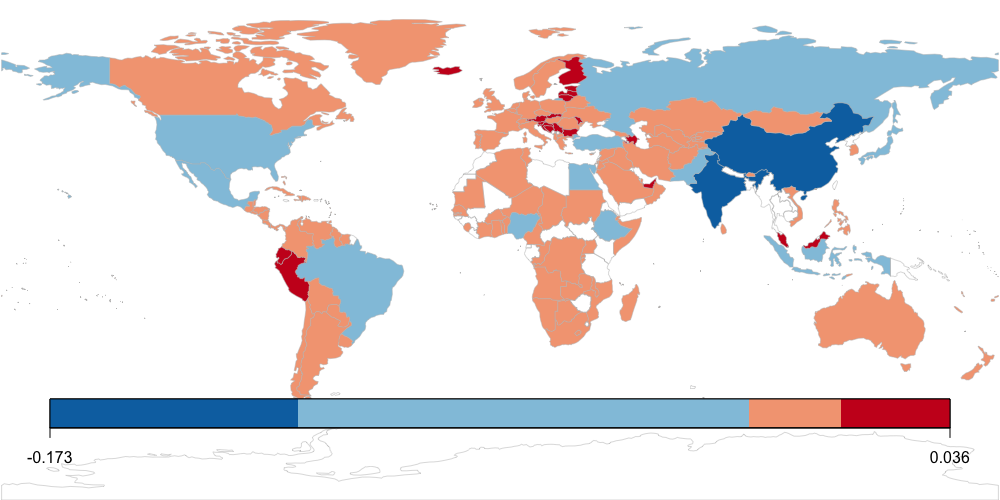}
	\caption{Representation of Web-Crawled Data (Red = over-represented i.r.t population)}
\end{figure}

Second, GDP-weighting and internet-usage-weighting have different effects on the two datasets. When adjusted for GDP, the correlation between Twitter data and population raises from $r=0.39$ to $r=0.60$, a fairly strong relationship and higher than when adjusted by internet usage. But when population is adjusted for GDP, the correlation between web-crawled data and population lowers from $r=0.39$ to $r=0.28$. In other words, economic information does not help to weight the web data towards actual ground-truth population density.

For web-crawled data, internet access provides a much better population weighting ($r=0.49$). This is perhaps not surprising because the internet usage statistics are directly relevant to the production of websites. But it is surprising that general internet access is not a good predictor of Twitter usage. Overall, we see that there is a definite relationship between populations and the amount of digital text produced per country, but there are clear regional biases.

What countries are specifically over-represented and under-represented in the two datasets? We first find the relative share of each dataset for each country. For example, what percentage of the web-corpus is from India? This assumes the total world population and the total corpus size as constants and finds the share of that total from each country. We then subtract the population estimates from the corpus-based estimates. In other words, we first normalize each representation (corpus size and population) and then find the difference between the normalized measures. This allows us to take into account the very different counts (words vs. persons).

If the result is negative, then a particular country is \textit{under-represented}. For example, the share of the corpus from India has values of -0.1730 (CC) and  -0.1421 (TW). This means that language from India is under-represented given what we would expect its population to produce. On the other hand, if the result is positive, then a particular country is \textit{over-represented}. For example, Estonia is over-represented in the web-crawled data (0.0290) as is Australia in the Twitter data (0.0226) These numbers mean that there is 2.9\% more language data from Estonia on the web than expected given the population of Estonia; and there is 17.3\% less language data from India on the web than expected given the population of India.

Countries are shown by their representation in Twitter (Figure 2) and the web corpus (Figure 3), with red indicating over-representation: there is more corpus data than population size would predict. The imbalance between Twitter data and population is caused by a clear over-representation of the US, Canada, western Europe, Russia, and South America. But the imbalance between web-crawled data and population has a very different geographic pattern: there is less extreme over-representation but more under-representation. Specifically, under-representation is apparent in Africa and Southeast Asia. 

The distribution of language data in Figures 2 and 3 raises an important distinction between types of users: locals vs. non-locals. For example, from internet usage statistics we know that many countries in Africa have less access to web-sites and thus produce much less web-crawled data. This is reflected in Figure 3. But many African countries are over-represented in Twitter data. Are these Twitter users members of the local populations or do they represent visitors?  Note that Figure 2 does not reflect the popularity of Twitter as a platform because we start by normalizing the Twitter output for each country against the total Twitter usage. The over-represented countries in Figure 2, then, represent places where Twitter data is produced at a higher rate than expected. It has nothing to do with the relative popularity of the platform (e.g., Twitter vs. web pages).

\section{Population Demographics}

\begin{figure}
    \centering
	\includegraphics[width=1\textwidth]{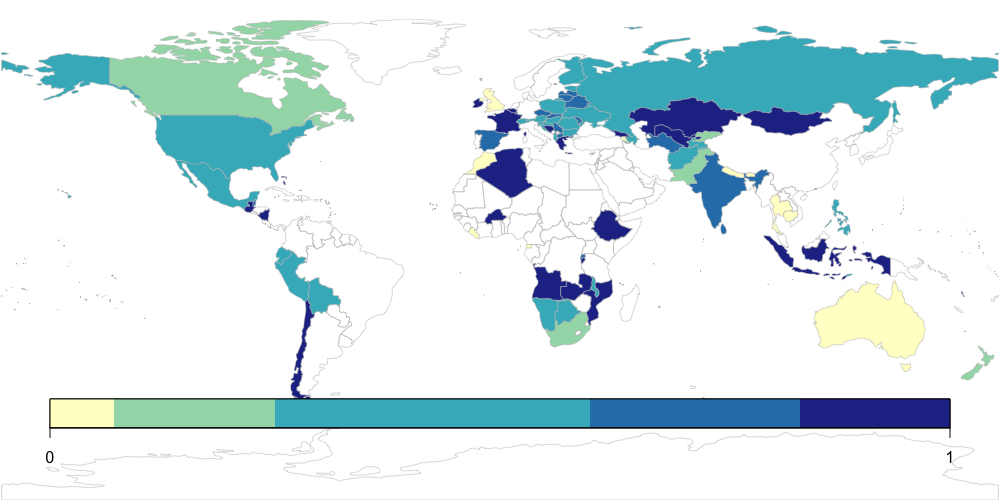}
	\caption{Web-Crawled Language Inventory in Relation to Ground-Truth}
	\includegraphics[width=1\textwidth]{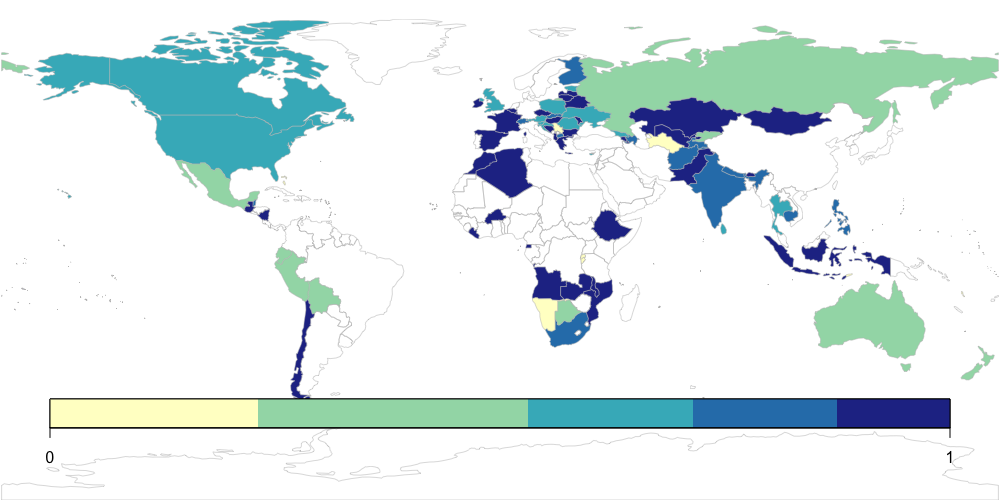}
	\caption{Twitter Language Inventory in Relation to Ground-Truth}
\end{figure}

\textbf{Can we use georeferenced corpora to determine what languages are used in a particular country?} We use as ground-truth the UN aggregated census data for each country and, in countries where this is not available, fall back on the CIA \textit{World Factbook} data. Instead of trying to match up exactly how much of the population uses a specific language, we instead say that a language is used in a country if at least 5\% of the observation is in that language. For example, if over 5\% of the population in India uses Hindi, then we expect to see Hindi make up at least 5\% of the corpora from India. This threshold allows us to evaluate the corpora without expecting that they will predict precisely the estimated figures of speakers per language. If there are ten languages in India that are used by over 5\% of the population, then we expect all ten languages to be present in the corpora from India.

Figures 4 and 5 show the true positive rate: what percent of the actual census-based languages for each country are found using text data? Figures 6 and 7, on the other hand, show the false positive rate: how many languages do the text datasets say are used in a country but are not found in census data? These are two simple methods for comparing the relationship between the corpora and the underlying populations. If we predicted that any language that makes up at least 5\% of the corpus from a country is, in fact, used by the population of that country, how often would be correct? There are many countries for which these two ground-truth datasets have no language information. For example, the UN language dataset has no information for Brazil. The ground-truth for language-use is much more sparse than that for population because many countries have no recent and reliable ground-truth data for the languages used by their populations. This lack of ground-truth data is not a problem. Rather, it is the motivation: if we can correctly predict language use in countries where we do have ground-truth, then we can use these methods to study countries that are currently unrepresented.

In both Figures 4 and 5 a darker red indicates a higher number of languages from a census being found in the respective corpora. In many cases, the two corpora agree in which languages they predict to be used in each country: Europe, those parts of Africa for which there is ground-truth data, and South America. But Twitter provides a better representation of North America and Oceania. One factor that is disguised in these figures is that many countries have only a few languages, so that a high true positive rate for a country could reflect only one or two languages. For example, both English and Spanish are very common on Twitter (c.f. Table 3), so that any country which predominantly uses these two languages will have a good representation by default. 

We can also think about the false positive rate: what languages do the corpora find that are not contained in the census-based ground-truth? For example, if English and Spanish are used in a country on Twitter but not reflected on the census, this is a false positive. As shown in Figure 6, the web-crawled corpus has very few false positive outside of Russia and eastern Europe. The situation on Twitter is similar: most false positives are in Russia and Eastern Europe, but in Twitter there are also over-predicted languages in the US and Canada, South Africa, France, India, and Australia. This is important because it shows that relying on Twitter alone would indicate that there are more diverse language speaking populations in these countries. As shown in Table 1, Eastern Europe accounts for 2.4\% of the world's population but 27.4\% of the web corpus; this explains the region's general false positive rate. For Russia, on the other hand, which is not included in Eastern Europe, the false positive rate cannot be explained in reference to general over-representation. In this case, the false positives are other European languages: French, German, Spanish, Italian. More research is needed to distinguish between immigration, tourism, and business as alternate sources of false positive languages appearing in digital data sets.

\begin{figure}
    \centering
	\includegraphics[width=1\textwidth]{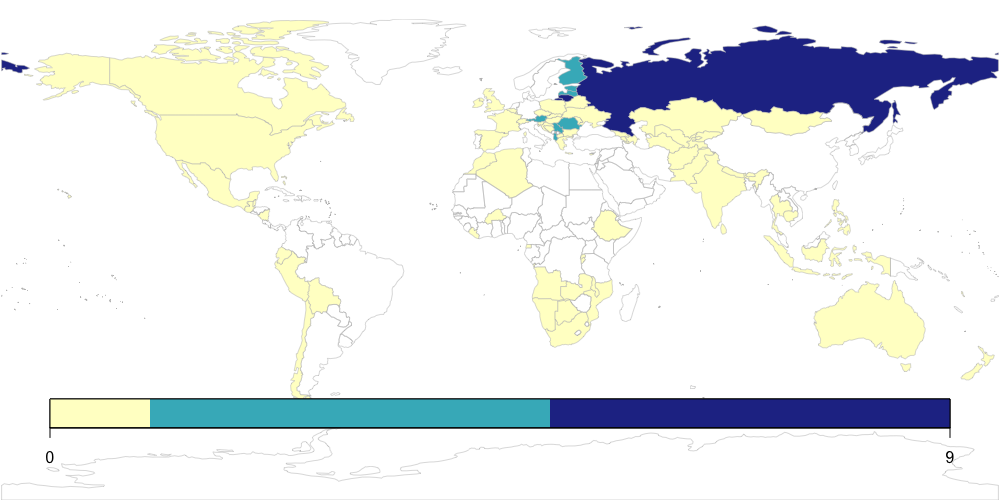}
	\caption{False Positive Languages By Country for Web-Crawled Corpus}
	\includegraphics[width=1\textwidth]{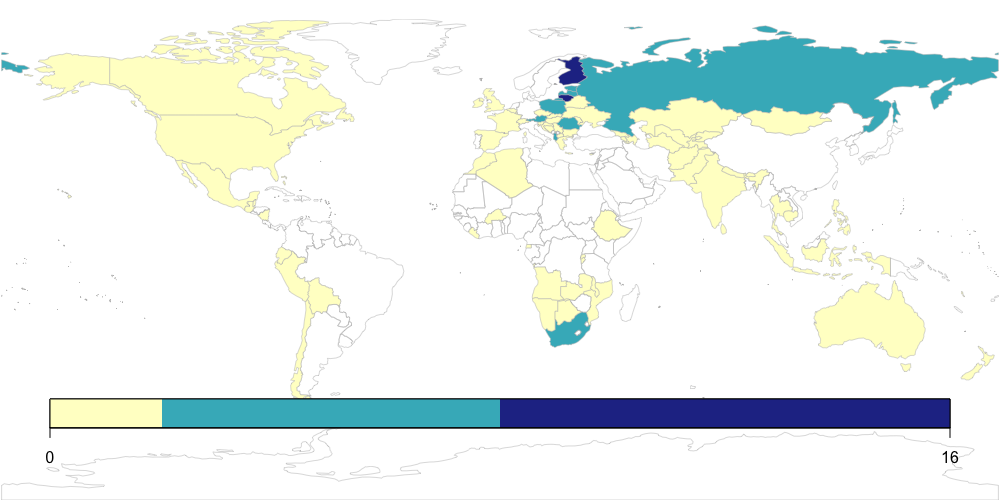}
	\caption{False Positive Languages By Country for Twitter Corpus}
\end{figure}

\section{Discussion}

Analyses and models based on digital texts, especially from Twitter, often come with uncertainty about the underlying populations that those texts represent. This paper has systematically collected Twitter and web-data from locations around the world without language-specific searches that would bias the collection. The purpose is to understand how well these data sets correspond with what we know about global populations from ground-truth sources, providing a method for evaluating different data collection techniques.

The first important finding is that patterns from Twitter and web-crawled data diverge significantly in their representation of the world's population. This simply reflects the fact that data drawn from Twitter and web pages will likely represent people from different places. Why? We have also seen that Twitter data matches populations better when population numbers are weighted by GDP and worse when weighted by internet-usage statistics. This implies that Twitter as a platform represents more wealthy populations than general web-crawled data. An alternate interpretation is that the Twitter collection here is based on urban areas, which tend to have more wealthy populations. Would the same bias be found with a rural-centered collection procedure? That is a secondary problem in this context because the goal is to develop ground-truth population-centered baselines that could be used to evaluate different Twitter collection methods.

The second important finding is that, given what ground-truth language-use data is available, there are in general very few \textit{false positives}: cases where the corpora suggest a language is frequently used in a country but census-based data does not. While uncommon, there are more false positives in Twitter data. This is significant because it means that, in general, these corpora do not predict language use that is not actually present.

But the third important finding is that, given what ground-truth language-use data is available, there remain a number of countries where these corpora do not represent all the language produced by the local populations: not all languages from censuses are found in digital texts. In this case Twitter has fewer missing languages.


\section{References}

\bibliography{geocomp}

\end{document}